\documentclass[letterpaper,10pt,journal,twoside]{IEEEtran}

\usepackage{amsmath,amsfonts}
\usepackage{algorithmic}
\usepackage{algorithm}
\usepackage{array}
\usepackage[caption=false,font=normalsize,labelfont=sf,textfont=sf]{subfig}
\usepackage{textcomp}
\usepackage{stfloats}
\usepackage{url}
\usepackage{verbatim}
\usepackage{graphicx}
\usepackage{booktabs}
\usepackage{makecell}
\usepackage{cite}
\usepackage{xcolor}
\usepackage{colortbl}

\title{\name: Real-Time World-Action Modeling for Agile UAV Navigation}

\author{
Runqing Wang, Ding Yu, Pengyuan Min, Xinhong Zhang, Wei Xiao, Yu Hu,\\ Jie Chen,~\IEEEmembership{Fellow,~IEEE}, Fu Zhang, \IEEEmembership{Senior Member,~IEEE}, and~Gang Wang,~\IEEEmembership{Senior Member,~IEEE}%
\thanks{Manuscript received: July 7, 2026; Revised: September 27, 2026; Accepted: November 2, 2026. This paper was recommended for publication by Editor Giuseppe Loianno upon evaluation of the Associate Editor and Reviewers' comments. This work was supported in part by the National Natural Science Foundation of China under Grants U23B2059 and 62688201, the Beijing Natural Science Foundation under Grant QY25271, and the Zhongguancun Academy under Grant 02012407. \emph{(Runqing Wang, Ding Yu, and Pengyuan Min contributed equally to this work.) (Corresponding author: Gang Wang.)}}
\thanks{Runqing Wang, Ding Yu, Pengyuan Min, Xinhong Zhang, Wei Xiao, Jie Chen, and Gang Wang are with the School of Automation, Beijing Institute of Technology, Beijing 100081, China (e-mail: bitwrq@bit.edu.cn; yd@bit.edu.cn; bitmpy@bit.edu.cn; xhzhang@bit.edu.cn; xiaowei@bit.edu.cn; chenjie@bit.edu.cn; gangwang@bit.edu.cn). Pengyuan Min and Xinhong Zhang are also with the Zhongguancun Academy, Beijing 100094, China.}
\thanks{Yu Hu is with Spatial Robotics Limited, Qianhai Shenzhen-Hong E-hub Northern District, Shenzhen 518052, China (e-mail: henry.hu@sparo.hk).}
\thanks{Fu Zhang is with the Department of Mechanical Engineering, The University of Hong Kong, Hong Kong (e-mail: fuzhang@hku.hk).}
\thanks{For supplementary video see: \protect\url{https://youtu.be/1EOsVur6vIE}. The code will be released at \protect\url{https://github.com/flyingbitac/FlowPilot}.}
\thanks{Digital Object Identifier (DOI): see top of this page.}
}

\begin{document}
\newcommand{\name}{FlowPilot}

\allowdisplaybreaks

\maketitle

\begin{abstract}

We present \name{}, the first world-action model for real-time onboard UAV navigation from depth. Unlike map-then-optimize pipelines that require local reconstruction or end-to-end policies that lack explicit scene prediction, \name{} jointly denoises future depth observations and executable trajectories with flow matching. A dual-stream mixture-of-transformers couples video and action experts through shared attention, allowing future-scene prediction and trajectory generation to inform each other. At deployment, the model runs action-centrically and outputs only a trajectory. To generate smooth, state-consistent references, actions are parameterized as degree-$7$ Bernstein polynomials: the current state constrains the initial control points, and the network predicts five free control points, yielding $C^2$-continuous references with closed-form velocity, acceleration and jerk. \name{} is trained on a three-level depth pyramid spanning high-throughput simulation, photorealistic simulation, and real onboard data. In closed-loop simulation, it outperforms learning- and optimization-based baselines under increasing clutter and commanded speeds up to $10$\,m/s. On a physical quadrotor, FlowPilot achieves a mean inference latency of $10.2$\,ms on a Jetson Orin NX and reaches a peak speed of $10$\,m/s during outdoor flight using only onboard sensing and computation.
\end{abstract}


\section{Introduction}

\IEEEPARstart{A}{utonomous} flight through unknown clutter is a core capability for search and rescue, inspection, and last-mile delivery~\cite{zhou2021egoplanner,loquercio2021agile}. The difficulty is not perception or planning alone, but their combination under the timing and tracking constraints of a small aerial robot. A quadrotor must convert onboard depth and state feedback into a reference trajectory within a few milliseconds, and that reference must be smooth enough for a high-rate controller to track with velocity, acceleration, and jerk feedforward terms.

Existing systems approach this problem from two directions. Map-then-optimize pipelines build a local map and solve a trajectory optimization problem~\cite{zhou2021egoplanner,wang2022gcopter,ren2025super}. They provide strong geometric structure and smooth trajectories, but every plan is computed from geometry that has already been reconstructed; the resulting latency, map noise, and accumulated perception errors become increasingly important as speed and clutter increase. Learning-based policies instead regress motion directly from observations~\cite{loquercio2021agile,lu2023yopo,yang2023iplanner,sridhar2024nomad}. They reduce perception-to-action latency and can tolerate imperfect sensing, but many such policies remain reactive: the action is predicted from the current observation without explicitly co-training a model of future scene evolution. Thus, neither paradigm directly provides a compact onboard model that couples anticipated future geometry with an immediately executable trajectory.

World-action models (WAMs) offer a natural way to introduce this coupling by learning future observations and actions in a generative model. In manipulation domains, future-video prediction provides dense supervision for how the world evolves, and jointly training video and action streams can yield policies that generalize better than action-only predictors~\cite{dreamzero2026,motus2025}. Recent results further suggest that the control benefit comes largely from co-training rather than from expensive test-time video generation, enabling action-centric inference~\cite{fastwam2026,gigaworld2026}. Yet, these models are designed for tabletop manipulation, where large backbones output low-dimensional action chunks at modest rates. Agile UAV navigation imposes a different set of critical constraints: the model must run on an embedded processor, the output must be trackable by a flight controller, and consecutive replans must remain temporally consistent.

\begin{figure*}[t]
    \centering
    \includegraphics[width=0.95\textwidth,height=9cm]{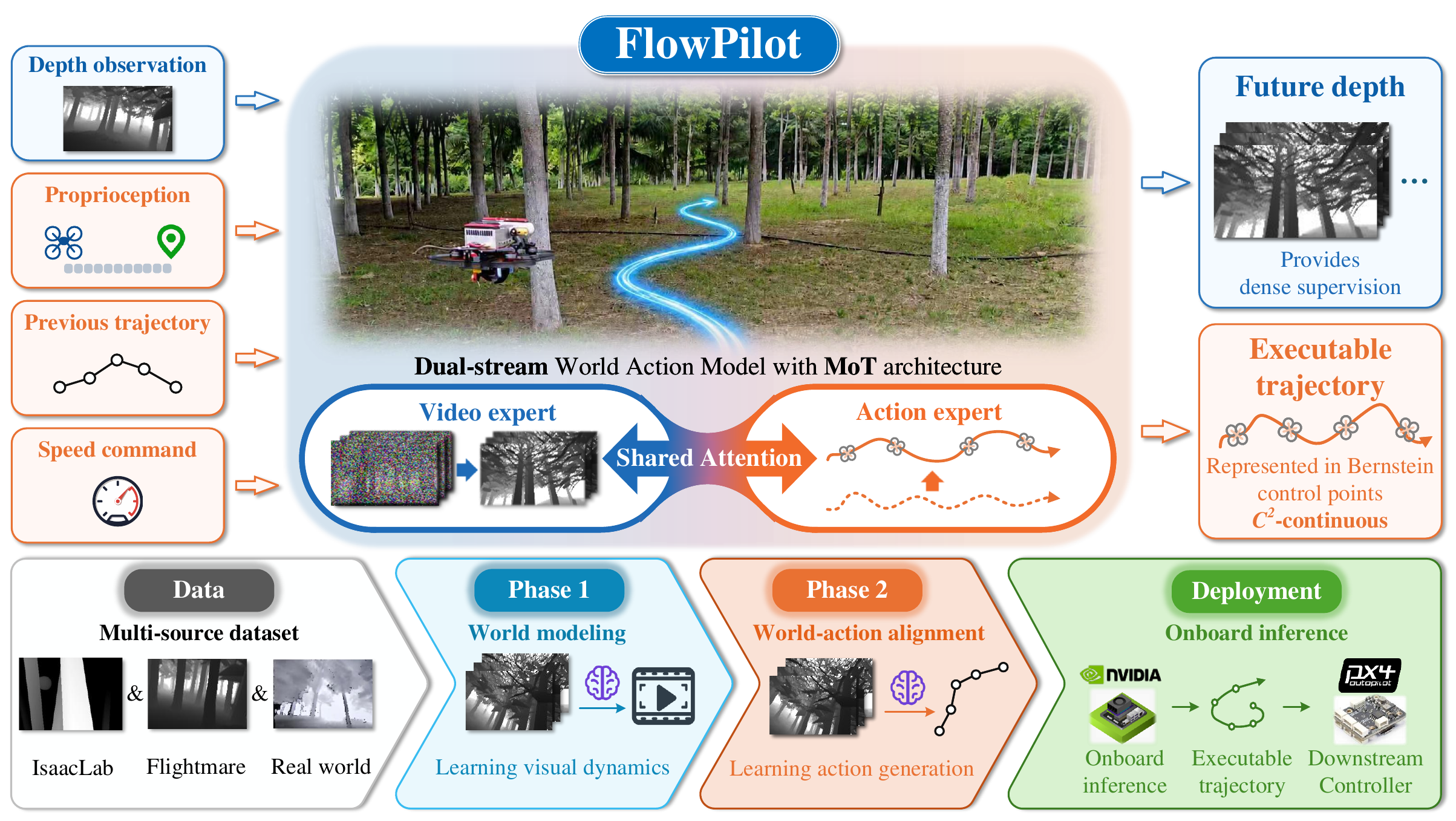}
    \caption{\textbf{Overview of FlowPilot.} FlowPilot receives the current depth observation, state-goal information, commanded speed, and previous trajectory in the current body frame. A dual-stream world-action model jointly trains future-depth prediction and Bernstein-trajectory generation through shared video-action attention. Training uses a three-level depth pyramid and proceeds from world-modeling pretraining to world-action alignment. At deployment, FlowPilot runs action-centrically onboard and sends smooth reference trajectories to a 100 Hz tracking controller.}
    \label{fig:teaser}
\end{figure*}

We propose FlowPilot, a lightweight depth-based WAM for real-time onboard UAV navigation (see Fig.~\ref{fig:teaser}). FlowPilot couples a depth-video expert and an action expert in a dual-stream mixture-of-transformers (MoT). During training, the two streams are jointly denoised with flow matching so that future-depth prediction and trajectory generation condition on each other through shared attention. During deployment, FlowPilot runs action-centrically: the controller consumes the generated trajectory, while future-depth decoding is not required in the control loop.

The action representation is critical. Rather than predicting independent waypoints, FlowPilot predicts the free control points of a degree-$7$ Bernstein polynomial whose initial control points are constrained by the current position, velocity and acceleration. The generated reference is thus smooth, state-consistent, and differentiable in closed form, while the denoised action latent remains small enough for real-time onboard inference. To reduce oscillation between replans, the model also receives previous trajectory as a soft conditioning token rather than warm-starting the denoising process from the last command. Our contributions are as follows.
\begin{itemize}
    \item \textbf{A compact WAM for onboard UAV navigation.} FlowPilot brings flow-matching WAM to real-time depth-based UAV navigation, coupling future-depth prediction and trajectory generation in a single dual-stream network.
    \item \textbf{A state-constrained generative action space.}
    FlowPilot predicts only the free control points of a
    state constrained Bernstein polynomial, yielding
    $C^2$-continuous trajectories while matching the generative
    action dimension to the trajectory's remaining degrees of freedom.
    \item \textbf{Closed-loop onboard validation.} FlowPilot runs fully onboard a Jetson Orin NX with a mean inference latency of $10.2$\,ms and is evaluated against optimization- and learning-based baselines in simulation and indoor and forest flights, reaching a peak speed of $10$\,m/s outdoors.
\end{itemize}

\section{Related Work}

\subsection{Mapping-and-Planning UAV Navigation}
Vision-based UAV navigation builds a local representation of free space and then optimizes a smooth trajectory over that representation. Systems such as EGO-Planner, MINCO-style planners, safety-oriented high-speed planners, and time-optimal planners remain strong references because they expose geometric structure and produce controller-friendly trajectories~\cite{zhou2021egoplanner,wang2022gcopter,ren2025super,qin2026perception}. Their limitation is structural: planning quality depends on the latency and accuracy of the reconstructed map. At high speed or in clutter, the vehicle can be forced to plan from incomplete or delayed geometry.

\subsection{Learning-Based Visual Navigation}
Learning-based UAV navigation reduces latency by predicting motion directly from depth, images, or learned state representations. Agile flight policies, end-to-end depth-based obstacle-avoidance networks, one-shot planners such as YOPO, and goal-conditioned visual policies have shown that learned controllers and planners can fly quickly and tolerate imperfect sensing~\cite{loquercio2021agile,icra225vit,lu2023yopo,yang2023iplanner,sridhar2024nomad,zhang2026mad}. However, such methods are primarily reactive and output waypoints or motion primitives without explicitly co-training a model of future observations. Related learning-augmented control studies improve robustness by
correcting learned dynamics or disturbances online~\cite{yang2026metadisturbance}. These methods address model uncertainty and tracking, whereas our focus is to generate collision-aware reference trajectories from onboard depth.

\subsection{World (Action) Models}
World models learn predictive representations of environment dynamics for planning and control~\cite{hafner2025dreamerv3,Zhang2024STORM}. They have been applied to vision-based drone flight through model-based reinforcement learning~\cite{romero2025dream}. WAMs extend this idea by generating actions jointly with predicted future observations, so the policy is grounded in an evolving world representation rather than learned as a separate head. Recent WAMs employ diffusion or flow-style objectives and MoT architectures to couple video and action streams~\cite{dreamzero2026,motus2025,bao2023unidiffuser}. Efficiency-oriented work further shows that future-video co-training can improve control even when future frames are not decoded at test time~\cite{fastwam2026,gigaworld2026}. Existing WAMs mainly target manipulation with large backbones and low-frequency action chunks. Recent work on aerial world models has addressed flight through long-horizon visual generation and navigation in 3D space~\cite{aerialwm2025}. However, this line of work does not focus on real-time onboard trajectory generation for agile quadrotors.

\subsection{Trajectory Representations for Learned Planners}
The action representation determines whether a learned plan is merely accurate in position or actually trackable. Independent waypoint regression is simple to supervise, but it provides no smoothness coupling between neighboring points and can yield large derivative oscillations after numerical differentiation. Optimization-based planners avoid this problem with smooth polynomial, B-spline, or MINCO representations that provide analytic derivatives~\cite{mellinger2011minimum,zhou2021egoplanner}. Recent residual-learning approaches also optimize control-friendly trajectory quality by learning corrections from execution feedback~\cite{guo2026optimizing}. Diffusion and flow policies, on the other hand, are attractive because they can model multi-modal trajectory distributions~\cite{chi2023diffusion,lipman2023flowmatching}.

\section{Method}

\subsection{Problem Formulation}
\label{subsec:problem}

We consider goal-directed navigation from a single onboard depth camera.
At each control instant, the vehicle observes a depth image $\mathbf{o}\in\mathbb R^{H\times W}$ and proprioceptive state and outputs a short-horizon reference trajectory for a downstream controller while replanning at a high rate.

All observations, states, goals, and trajectories are expressed in a per-sample body frame anchored at the drone's current position and heading,
so the network reasons in ego-centric rather than global geometry.
We stack the proprioceptive state and navigation goal into a single state-goal vector
\begin{equation}
    \label{eq:state}
\mathbf{s}=[\mathbf{p}_0,\;\mathbf{v}_0,\;\mathbf{a}_0,\;\mathbf{j}_0,\;\mathbf{q},\;\mathbf{p}_g,\;\mathbf{v}_g,\;\mathbf{a}_g]\in\mathbb{R}^{25},
\end{equation}
where $\mathbf{p}_0,\mathbf{v}_0,\mathbf{a}_0,\mathbf{j}_0\in\mathbb{R}^3$ are the current position, velocity, acceleration, and jerk, $\mathbf{q}\in\mathbb{R}^4$ is the orientation quaternion, and $\mathbf{p}_g,\mathbf{v}_g,\mathbf{a}_g\in\mathbb{R}^3$ denotes the goal position, velocity, and acceleration, respectively, all in the body frame.

FlowPilot, parameterized by $\theta$, maps the current depth observation $\mathbf{o}$, state-goal vector $\mathbf{s}$, commanded cruise speed $c\in\mathbb R$, and predicted trajectory $\tau_{\mathrm{prev}}$ to two outputs
$f_\theta(\mathbf{o},\mathbf{s},c,\tau_{\mathrm{prev}})
    \rightarrow
(\hat{\mathbf{z}}^{\mathrm{V}},\hat{\mathbf{z}}^{\mathrm{A}})$,
where $\hat{\mathbf{z}}^{\mathrm{V}}$ is the future-depth latent, and $\hat{\mathbf{z}}^{\mathrm{A}}\in\mathbb{R}^{5\times3}$ denotes the five free control points of the action trajectory in normalized Bernstein space.
The trajectory is represented by the free control points of a state-constrained Bernstein polynomial (Sec.~\ref{subsec:bernstein}), and both outputs are generated jointly in a single flow-matching pass (Sec.~\ref{subsec:fm}) so that the predicted geometry and motion condition on each other.

\subsection{Network Architecture}
\label{subsec:mot}

\begin{figure}[t]
    \centering
    \includegraphics[width=\columnwidth]{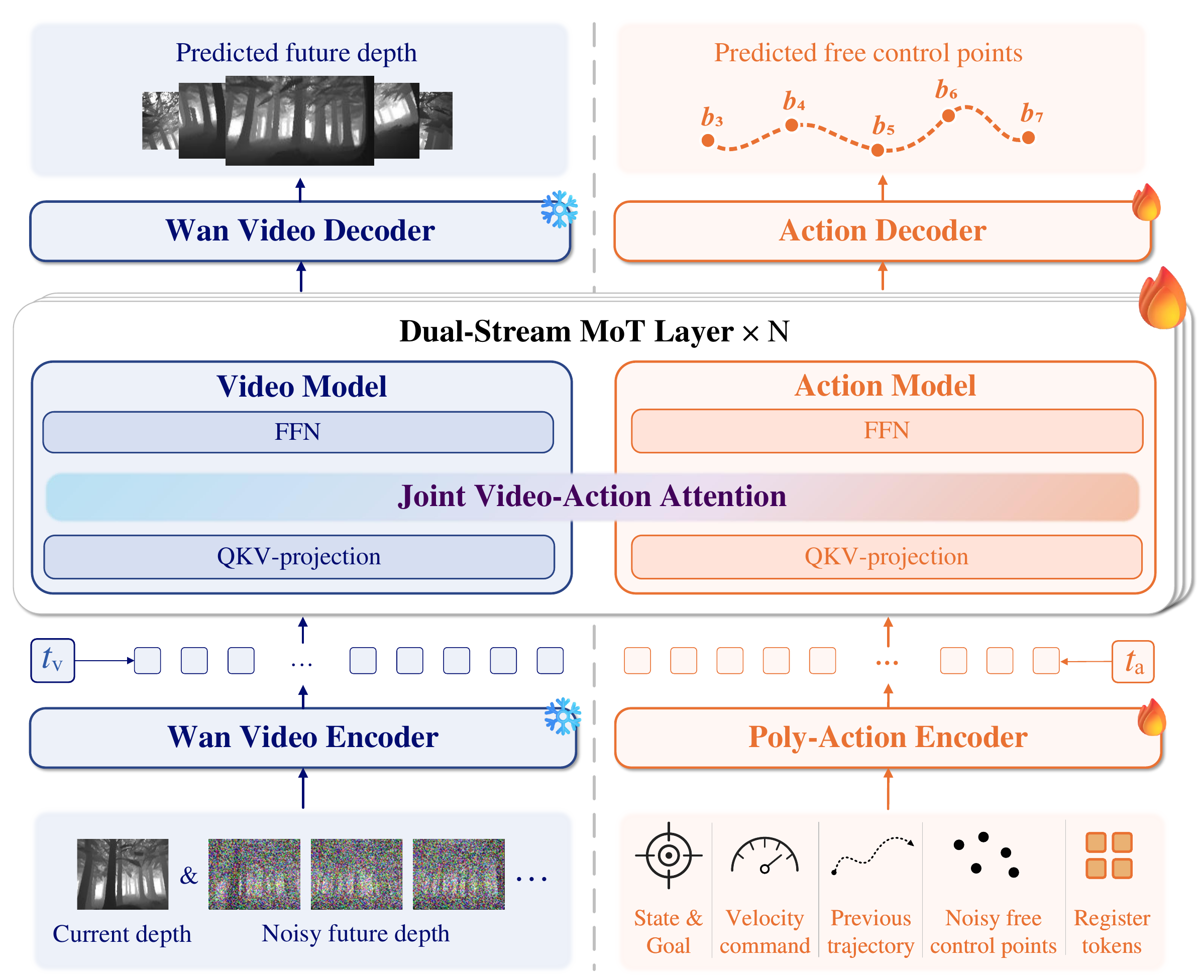}
    \caption{\textbf{Dual-stream MoT architecture.} The video stream predicts future depth and the action stream generates Bernstein control points, while the two streams exchange information through shared video-action attention.}
    \label{fig:mot_architecture}
\end{figure}

FlowPilot couples world modeling and action generation in a single MoT (Fig.~\ref{fig:mot_architecture}), where video and action experts keep modality-specific weights but exchange information through joint self-attention.
The future-geometry and motion representations are denoised together during training, while deployment remains action-centric.

\paragraph{Video expert} models how the scene evolves.
It takes the depth observation-one conditioning frame and eight future frames-and encodes it with a frozen Wan2.2 VAE~\cite{wan2025} into latent tokens that carry rotary 3D (RoPE-3D \cite{rope}) position encodings.
From the noisy future-depth latent the expert predicts a velocity field in the VAE latent space, i.e., the denoising direction of the future depth.

\paragraph{Action expert} produces the motion.
A state-action encoder embeds the state-goal vector $\mathbf{s}$ in \eqref{eq:state}, the commanded cruise speed $c$, the noisy action latent, and a previous-trajectory token.
From these tokens, the expert predicts an action-latent velocity field that denoises the free Bernstein control points of the trajectory (Sec.~\ref{subsec:bernstein}).

\paragraph{Shared attention} couples the two streams.
In every MoT layer, both streams are modulated by their flow-matching times ($t_v$ for video, $t_a$ for action), projected into a shared space, attended jointly, and split back to modality-specific feed-forward paths.
Through this shared attention, the action tokens attend to the predicted future geometry while the video tokens attend to the action being denoised within the same step.
The future-depth prediction is thus conditioned on the action being denoised, not merely on the current frame, state and goal; conversely, the action stream can attend to a representation trained to predict future geometry.
This mutual conditioning makes FlowPilot a world \emph{action} model rather than a goal-conditioned visual prior with a separate policy head.
Keeping modality-specific weights and feed-forward paths lets each stream preserve its own representation while sharing information at attention layers.

\paragraph{Action-centric deployment}
uses the jointly trained model without requiring future-depth decoding in the control loop.
During training, the video stream supplies dense supervision for scene evolution and couples to the action stream through shared attention.
During onboard flight, the controller consumes only the denoised Bernstein control points; future-depth tokens may be decoded for diagnostics, but are not needed to issue the next reference.
This design lets the policy benefit from world-model co-training while keeping the deployed perception-to-action path compact.

\subsection{Bernstein Polynomial Action Representation}
\label{subsec:bernstein}

\begin{figure}[t]
    \centering
    \includegraphics[width=0.95\columnwidth]{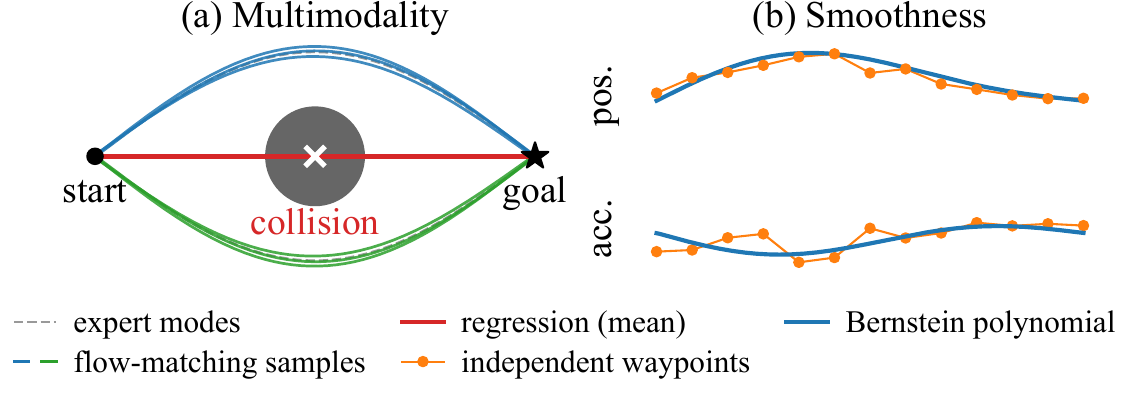}
    \caption{\textbf{Trajectory-generation motivation.} \textbf{(a)} Flow matching preserves multiple feasible modes. \textbf{(b)} Bernstein control points provide smooth trajectories with analytic derivatives, unlike independent waypoint predictions.}
    \label{fig:motivation}
\end{figure}

The action representation influences the smoothness and tracking performance of generated trajectories.
A sequence of independent waypoints is easy to supervise, but its weakness is hidden by position-error metrics: a waypoint mean absolute error (MAE) measures only \emph{zeroth-order} accuracy, whereas a flight controller also tracks the trajectory's velocity, acceleration and jerk as feedforward terms.
Without explicit smoothness constraints, errors in predicted waypoints can be amplified by numerical differentiation, degrading tracking performance. Figure~\ref{fig:motivation}(b) provides a schematic illustration.

We choose a degree-seven polynomial to match the degree used by SUPER~\cite{ren2025super}, our expert trajectory generator.
We represent the action not as waypoints but as a smooth curve with closed-form derivatives: a degree-$7$ Bernstein polynomial in $\mathbb{R}^3$ over the horizon $T$ with $8$ control points $\mathbf{b}_k\in\mathbb{R}^3$
\begin{equation*}
    \mathbf{p}(t)=\sum_{k=0}^{7}\mathbf{b}_k\,B_{k,7}(t/T),
    \quad
    B_{k,7}(u)=\binom{7}{k}u^{k}(1-u)^{7-k}.
\end{equation*}
The first three control points are hard-constrained from the current state so that the trajectory matches the drone's position $\mathbf{p}_0$, velocity $\mathbf{v}_0$, and acceleration $\mathbf{a}_0$ at $t=0$
\begin{equation}
\mathbf{b}_0=\mathbf{p}_0,\quad
    \mathbf{b}_1=\mathbf{b}_0+\tfrac{T}{7}\mathbf{v}_0,\quad
    \mathbf{b}_2=2\mathbf{b}_1-\mathbf{b}_0+\tfrac{T^2}{42}\mathbf{a}_0 .
\end{equation}
Fixing $\mathbf{b}_0,\mathbf{b}_1,\mathbf{b}_2$ consumes three of the eight control points, leaving exactly five free ones, so the action expert predicts only $\mathbf{b}_3,\dots,\mathbf{b}_7$ (in a normalized space).
Stacking these five points row-wise gives the $5\times3$ action latent that flow matching denoises; its $15$ degrees of freedom are precisely the trajectory's remaining degrees of freedom, so the action dimension and the polynomial are matched by construction.
Differentiating the Bernstein form yields velocity, acceleration, and jerk again as Bernstein polynomials of the differenced control points
\begin{align*}
    \dot{\mathbf{p}}(t)&=\tfrac{7}{T}\textstyle\sum_{k=0}^{6}(\mathbf{b}_{k+1}-\mathbf{b}_k)B_{k,6}(t/T),\\
    \ddot{\mathbf{p}}(t)&=\tfrac{42}{T^2}\textstyle\sum_{k=0}^{5}(\mathbf{b}_{k+2}-2\mathbf{b}_{k+1}+\mathbf{b}_k)B_{k,5}(t/T),\\
    \dddot{\mathbf{p}}(t)&=\tfrac{210}{T^3}\!\textstyle\sum_{k=0}^{4}(\mathbf{b}_{k+3}-3\mathbf{b}_{k+2}+3\mathbf{b}_{k+1}-\mathbf{b}_k)B_{k,4}(t/T),
\end{align*}
so the full kinematic state is available in closed form.

The Bernstein representation provides analytic derivatives, avoiding numerical differentiation of predicted waypoints. The hard constraints ensure the reference trajectory matches the measured position, velocity, and acceleration at its start.
Dynamic feasibility still depends on the downstream tracking controller and vehicle limits; the role of Bernstein representation is to provide a smooth, state-consistent reference with analytic derivatives.
Predicting five control points instead of waypoints also shrinks the action stream from tens of tokens to five, reducing the joint-attention length and inference cost.

\subsection{Training Data: A Three-Level Depth Pyramid}
\label{subsec:data}

Data-pyramid strategies in robot learning combine scalable, broadly informative data with more targeted robot experience, as exemplified by GR00T N1~\cite{grootn12025}.
For depth-based UAV navigation, we organize training observations into three levels according to collection scalability and relevance to onboard sensing.
The base consists of $16$\,h of depth from a GPU-parallel, high-throughput IsaacLab simulation~\cite{diffaero}, which supplies the raw scale needed to cover diverse geometry and motion.
The middle contains $8$\,h of photorealistic and geometrically complex depth rendered in Flightmare, adding visual and structural realism that the faster simulator omits.
The apex contains $2$\,h of depth collected on a physical quadrotor, which carries the true onboard sensor statistics including noise, dropouts, and range limits that the model must face at deployment.
This organization combines broad geometric and motion coverage from simulation with sensor characteristics captured during real flight.
The resulting data pyramid is depicted together with the two-phase training masks in Fig.~\ref{fig:data_pyramid_attention}.

\subsection{Flow-Matching Objective}
\label{subsec:fm}

We model a trajectory distribution to preserve distinct valid maneuvers. For example, when either side of an obstacle is traversable, a deterministic regressor may average the two modes into an infeasible trajectory, whereas a generative model can sample either mode, as depicted in Fig.~\ref{fig:motivation}(a).
Both streams are trained with linear-interpolation flow matching~\cite{lipman2023flowmatching}.
Given a clean latent $\mathbf{z}_0$ and noise $\boldsymbol{\epsilon}\sim\mathcal{N}(0,\mathbf{I})$, the noisy latent at noise level $\sigma\in[0,1]$ and its velocity target are
\begin{equation*}
    \mathbf{z}_\sigma=(1-\sigma)\mathbf{z}_0+\sigma\boldsymbol{\epsilon},
    \qquad
    \mathbf{u}=\boldsymbol{\epsilon}-\mathbf{z}_0 ,
\end{equation*}
and the network predicts the velocity field $\mathbf{v}_\theta=f_\theta(\mathbf{z}_\sigma,\sigma,\mathbf{c})$ from the conditioning $\mathbf{c}$, where $\theta$ denotes the network parameters introduced in Sec.~\ref{subsec:problem}.
We draw $\sigma$ from a shifted schedule that concentrates training on the high-noise regime
\begin{equation*}
    \tilde{\sigma}=\frac{s\,\sigma}{1+(s-1)\,\sigma},\qquad s=5 .
\end{equation*}
FlowPilot applies this objective to two latents within one forward pass: the normalized free control points $\mathbf{z}^{\mathrm{A}}_0$ and the future-depth VAE latent $\mathbf{z}^{\mathrm{V}}_0$.
Following UniDiffuser~\cite{bao2023unidiffuser}, the two streams use independent noise levels $\sigma_\mathrm{A},\sigma_\mathrm{V}$, and the model minimizes the sum of their velocity losses
\begin{align}
    \mathcal{L}^{\theta}_{\mathrm{action}}&=\mathbb{E}_{\substack{(\mathbf{o},\mathbf{s},c,\mathbf{z}^{\mathrm{A}}_0,\mathbf{z}^{\mathrm{V}}_0)\sim\mathcal{D}\\ \sigma_\mathrm{A}\sim\mathcal{U}(0,1),\,\boldsymbol{\epsilon}_\mathrm{A}\sim\mathcal{N}(\mathbf{0},\mathbf{I})}}\big\|\mathbf{v}^{\theta}_\mathrm{A}-(\boldsymbol{\epsilon}_\mathrm{A}-\mathbf{z}^{\mathrm{A}}_0)\big\|_2^2,\\
    \mathcal{L}^{\theta}_{\mathrm{video}}&=\mathbb{E}_{\substack{(\mathbf{o},\mathbf{s},c,\mathbf{z}^{\mathrm{A}}_0,\mathbf{z}^{\mathrm{V}}_0)\sim\mathcal{D}\\ \sigma_\mathrm{V}\sim\mathcal{U}(0,1),\,\boldsymbol{\epsilon}_\mathrm{V}\sim\mathcal{N}(\mathbf{0},\mathbf{I})}}\big\|\mathbf{v}^{\theta}_\mathrm{V}-(\boldsymbol{\epsilon}_\mathrm{V}-\mathbf{z}^{\mathrm{V}}_0)\big\|_2^2,\\
    \mathcal{L}^{\theta}&=\mathcal{L}^{\theta}_{\mathrm{action}}+\lambda\,\mathcal{L}^{\theta}_{\mathrm{video}},
\end{align}
where $\mathbf{v}^{\theta}_\mathrm{A}$ and $\mathbf{v}^{\theta}_\mathrm{V}$ are the two streams' predicted velocity fields. We set $\lambda=1$, assigning equal weights to the action and video losses. Because the two noise levels are sampled independently, training spans all $(\sigma_\mathrm{A},\sigma_\mathrm{V})$ pairs, including the world-model regime (clean action, noisy video) and the policy regime (clean video, noisy action). The joint behavior of Sec.~\ref{subsec:mot} is \emph{learned} rather than hand-designed, and a single network serves as both a world model and a policy.

\subsection{Two-Phase Training: A Navigation Prior, then World-Action Alignment}
\label{subsec:training}

\begin{figure}[t]
    \centering
    \includegraphics[width=0.95\columnwidth]{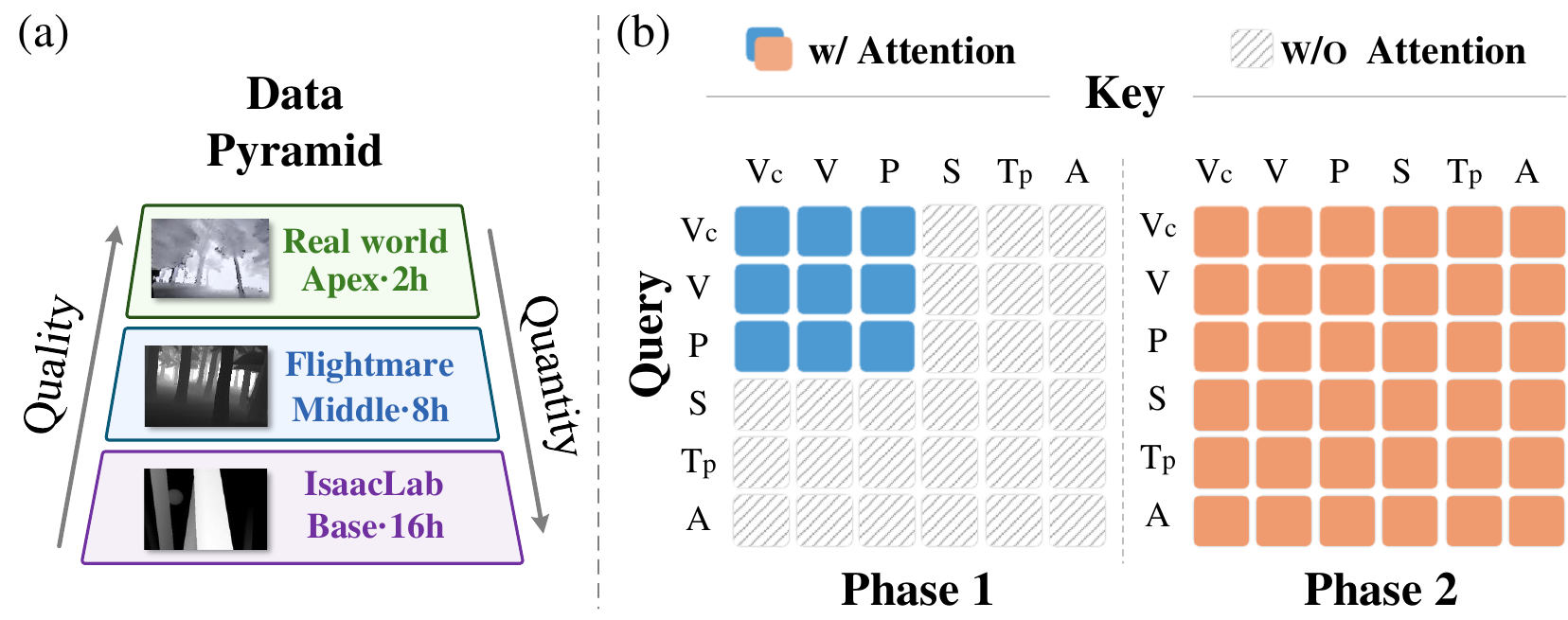}
    \caption{\textbf{Two-phase training.} Left: three-level depth pyramid. Right: attention masks for Phase~1 model pretraining and Phase~2 joint video-action training.}
    \label{fig:data_pyramid_attention}
\end{figure}

We train FlowPilot in two phases that separate \emph{learning how the world evolves} from \emph{learning how to act in it}.
The motivation mirrors that of foundation models: grounding a precise, high-quality policy on top of a model that already captures broad navigation dynamics is easier than learning both at once from scarce expert trajectories.
Fig.~\ref{fig:data_pyramid_attention} shows how the depth pyramid of Sec.~\ref{subsec:data} is used together with different token-level attention patterns in the two phases.

\paragraph{Phase 1: World-model pretraining}
We pretrain the video stream alone on the depth pyramid.
From the current observation and the state-goal conditioning token, the model learns to predict future depth frames, instilling a broad prior over how cluttered scenes evolve as a drone flies through them.
In Fig.~\ref{fig:data_pyramid_attention}, the attention mask in this stage is restricted to the world-modeling tokens: the conditioning-frame token $V_c$, future-video tokens $V$, and state-goal conditioning token $P$ attend to one another, while the speed-command token $S$, previous-trajectory token $T_p$, and action token $A$ are masked out.
This stage learns a broad navigation prior, which serves as the representation on which the action policy is later built.

\paragraph{Phase 2: World-action alignment}
This phase adds the action expert and trains the full dual-stream model jointly, specializing the navigation prior into a precise policy.
In contrast to the restricted attention used during pretraining, Phase~2 enables full joint attention over all tokens, so the future-depth representation and the action representation can condition on each other within the same denoising process.
Here the supervision quality matters, so the action targets are high-quality expert trajectories generated by SUPER~\cite{ren2025super}, a high-speed navigation system rather than by the optimizer used only for coarse coverage in Phase~1.
To keep this supervision both high-quality and diverse, we collect SUPER trajectories only in forested Flightmare environments.

In Phase~2, both losses of Sec.~\ref{subsec:fm} are active.
The video loss continues to sharpen the world model toward a higher-quality motion trend, while the action loss aligns the generated trajectory with the predicted future geometry.
Keeping the action loss on the predicted velocity of the normalized free control points, rather than expanding to waypoints and matching them, avoids amplifying the prediction error through the polynomial's derivatives.
Because the action is supervised in the same forward pass that predicts the future depth, the policy is grounded in the model's predicted geometry rather than learned in isolation.

\paragraph{Previous-trajectory conditioning for replan consistency}
FlowPilot replans continuously, and each inference is an independent draw from noise.
Two consecutive plans computed from nearly identical observations can differ, and this inter-replan inconsistency excites oscillation in the downstream controller.
A remedy is to \emph{warm-start} the denoising process from the previous plan, hard-anchoring each new trajectory to the last one.
In practice, this shortcut backfires: the model learns the trivial solution of reproducing its previous output, either collapsing the flow-matching stochasticity into a near-copy of the last plan or progressively shortening the trajectory toward a degenerate fixed point-both of which we observed when warm-starting at inference time.
We instead supply the previous plan as a \emph{soft} conditioning signal.
The action expert receives an additional token representing the previous trajectory through its eight Bernstein control points, expressed in the current body frame. Following the classifier-free training strategy~\cite{cfg}, we replace this token with a learned null embedding with probability $0.2$, enabling the model to operate both with and without previous-trajectory conditioning.
Acting as a soft regularizer rather than a hard constraint, the conditioning steers successive plans toward temporal consistency while preserving the multimodality of the flow-matching distribution and the full trajectory horizon.

\section{Experiments}

In this section, we demonstrate that FlowPilot generates robust, trackable trajectories for agile navigation from onboard depth. We first compare FlowPilot with optimization- and learning-based baselines in closed-loop simulation under increasing clutter and commanded speed. We then study the design choices affecting denoising, prediction horizon, and trajectory representation. Finally, we validate the onboard timing and real-world transfer of the same stack through indoor and outdoor flight experiments.

\begin{table*}[t]
    \centering
    \caption{Ablation of Design Choices.}
    \label{tab:design_ablation}
    \begingroup
    \footnotesize
    \resizebox{1\textwidth}{!}{\begin{tabular}{lllcc|cccccc}
    \toprule
    \multicolumn{5}{c|}{Design choices} & \multicolumn{6}{c}{Metrics} \\
    \midrule
    Architecture & \makecell{Denoising\\strategy} & \makecell{Action space} & \makecell{Denoising\\steps} & \makecell{Horizon (s)} & \makecell{Success\\rate (\%)} & \makecell{Mean\\clearance (m)} & \makecell{Avg. speed\\(m/s)} & \makecell{Max. speed\\(m/s)} & \makecell{Inference\\latency (ms)} & \makecell{Normalized jerk\\(s$^{-4}$)} \\
    \midrule
    Single & Both & Bernstein & 2 & 1.6 & $\boldsymbol{100}$ & $0.440$ & $4.317$ & $5.668$ & $11.480$ & $5.585$ \\
    MoT & Depth-Frozen & Bernstein & 2 & 1.6 & $80$ & $0.357$ & $4.043$ & $5.202$ & $-$ & $4.756$ \\
    \midrule
    MoT & Both & Waypoints & 2 & 1.6 & $10$ & $0.396$ & $4.445$ & $\boldsymbol{6.753}$ & $-$ & $15.395$ \\
    \midrule
    MoT & Both & Bernstein & 3 & 1.6 & $\boldsymbol{100}$ & $0.449$ & $4.765$ & $5.982$ & $15.364$ & $3.784$ \\
    MoT & Both & Bernstein & 4 & 1.6 & $\boldsymbol{100}$ & $0.385$ & $4.725$ & $5.958$ & $20.114$ & $4.672$ \\
    MoT & Both & Bernstein & 5 & 1.6 & $95$ & $0.375$ & $4.722$ & $6.094$ & $24.813$ & $5.430$ \\
    \midrule
    MoT & Both & Bernstein & 2 & 0.8 & $\boldsymbol{100}$ & $0.447$ & $4.684$ & $5.790$ & $\boldsymbol{9.549}$ & $\boldsymbol{3.734}$ \\
    MoT & Both & Bernstein & 2 & 2.4 & $95$ & $0.399$ & $\boldsymbol{4.965}$ & $6.040$ & $14.501$ & $3.804$ \\
    \midrule
    \rowcolor{gray!15}
    MoT & Both & Bernstein & 2 & 1.6 & $\boldsymbol{100}$ & $\boldsymbol{0.488}$ & $4.765$ & $6.159$ & $10.234$ & $4.075$ \\
    \bottomrule
    \end{tabular}}
    \endgroup
\end{table*}

\subsection{Experimental Setup}
\label{subsec:experimental_setup}

The experimental setup separates workstation-based evaluation from onboard execution. All simulation experiments, ablation studies, and baseline comparisons are conducted on a workstation equipped with an Intel Core i9-14900K CPU and an NVIDIA GeForce RTX 4090 GPU. Unless noted otherwise, FlowPilot decodes each action with two Euler denoising steps over a $1.6$-second horizon.
For the closed-loop ablations in Tables~\ref{tab:design_ablation} and~\ref{tab:tracking_ablation}, we fix the commanded speed at $v_{\mathrm{cmd}}=6$\,m/s and the obstacle sparsity at $15$, using identical scenes and start--goal pairs. Training and evaluation use independently generated random maps, with no map reuse between the two sets.
Success rate is the fraction of collision-free trials that reach the goal within $120$\,s.

\begin{figure}[!t]
    \centering
    \includegraphics[width=1\columnwidth]{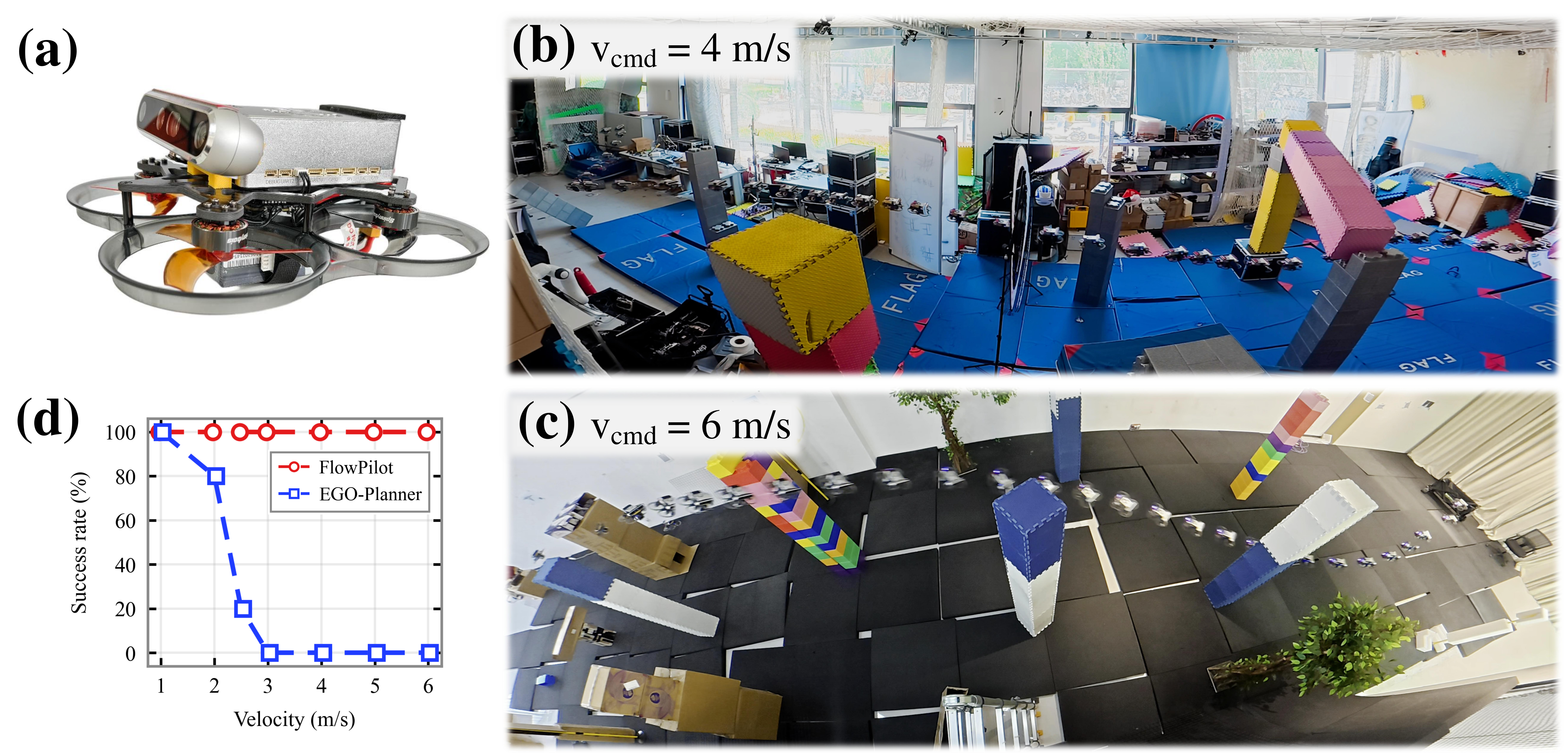}
    \caption{\textbf{Real UAV platform and indoor test environments.}
    \textbf{(a)} Onboard platform used for closed-loop flight. \textbf{(b)} A $4\times10$\,m$^2$ indoor area with $10$ obstacles (sparsity=$4.00$). \textbf{(c)} A $5\times10$\,m$^2$ indoor area with $13$ obstacles (sparsity=$3.85$); over $10$ trials: mean path length $1.29\times$ the straight-line distance, $3$--$4$ maneuvers per flight, mean minimum obstacle clearance $0.53$\,m, median free-corridor width $0.67$\,m. \textbf{(d)} Indoor success rate versus commanded velocity, evaluated in the layout of (c).}
    \label{fig:platform_indoor}
\end{figure}

All real-world experiments use the onboard platform in Fig.~\ref{fig:platform_indoor}. The quadrotor is built on an OddityRC 35Pro frame with T-Motor F60 PRO IV 2150\,KV motors, an NxtPX4v2 flight controller, a MicoAir 50A 4-in-1 electronic speed controller, and an NVIDIA Jetson Orin NX Super $16$\,GB module for onboard computation. An Intel RealSense D455 camera (maximum range $6$\,m) provides depth, resized to $160\times96$ pixels before processing. The entire stack runs onboard without ground-station computation or offline mapping: VINS-Fusion \cite{vins} estimates the state, FlowPilot replans in real time, and OMMPC\cite{ommpc} tracks the latest reference trajectory at $100$\,Hz.

\subsection{Simulation Experiments}
\label{subsec:simulation_experiment}

We evaluate closed-loop navigation in simulated forests with randomly distributed obstacles. Obstacle sparsity is defined as the area per obstacle, with smaller values indicating denser clutter; a sparsity of $10$ corresponds to one obstacle per $10$\,m$^2$. All methods are evaluated on the same scenes and start-goal pairs, using identical commanded speeds and success criteria. Each method must reach the goal without a prebuilt map: FlowPilot, FlowPilot (action-only), EGO-Planner, and YOPO operate on the same onboard depth and state, whereas SUPER~\cite{ren2025super} plans from LiDAR measurements and is included as a privileged-sensing expert reference. FlowPilot (action-only) uses the same Bernstein action representation but does not generate future depth or use shared video-action attention.

\begin{figure}[t]
    \centering
    \includegraphics[width=0.95\columnwidth]{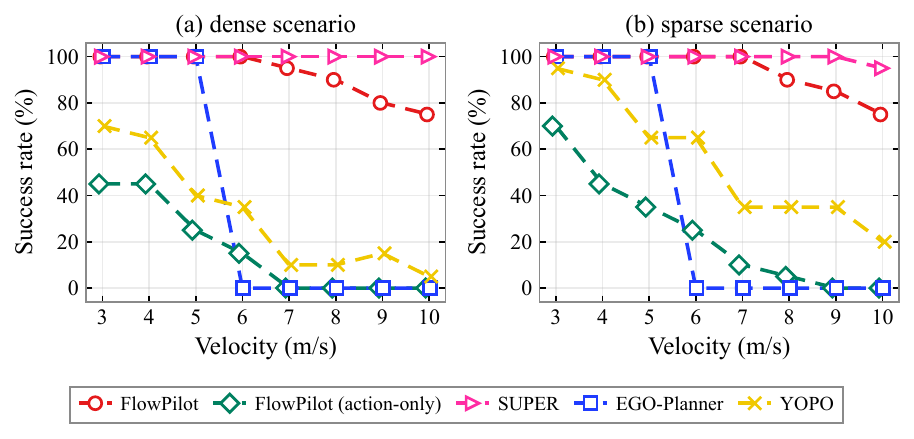}
    \caption{\textbf{Success rate under different obstacle sparsities and commanded velocities.}
    FlowPilot is compared with FlowPilot (action-only), SUPER, EGO-Planner, and YOPO over \textbf{20 trials} per velocity and sparsity level.
    \textbf{(a)} Sparsity $=10$.
    \textbf{(b)} Sparsity $=15$.}
    \label{fig:sim_success_rate}
\end{figure}

Across commanded speeds from $3$ to $10$\,m/s (Fig.~\ref{fig:sim_success_rate}), SUPER, the LiDAR-based expert that supervises Phase-2 training, remains near $100\%$ in both dense (sparsity $10$) and sparse (sparsity $15$) forests. Using only onboard depth, FlowPilot most closely tracks this expert: it stays at or above $90\%$ up to $8$\,m/s and degrades gently to about $75\%$ at $10$\,m/s, with the dense scenario consistently harder than the sparse one. In contrast, FlowPilot (action-only) reaches zero success by $9$\,m/s in both scenarios; EGO-Planner collapses to near zero beyond $5$\,m/s, while YOPO degrades steadily as speed or clutter increases. The margin between FlowPilot and the depth-based baselines is largest precisely where replanning margins are tightest--at high speed and in dense clutter--indicating that coupling predicted future geometry with action denoising is most beneficial under such conditions.

We ablate four design factors in the closed-loop setting: the denoising strategy and stream architecture, the trajectory representation, the number of Euler denoising steps, and the prediction horizon. The gray row in Table~\ref{tab:design_ablation} denotes the default FlowPilot configuration, which jointly denoises future depth and Bernstein actions with two Euler steps over a $1.6$-s horizon. Joint future-depth denoising and dual-stream attention provide a favorable trade-off between safety, speed, and smoothness. Bernstein offers a more favorable balance than waypoint reconstruction, while additional denoising steps increase latency without consistent closed-loop gains. Varying the horizon further exposes the performance-latency trade-off.

\begin{figure}[!t]
    \centering
    \includegraphics[width=0.95\columnwidth]{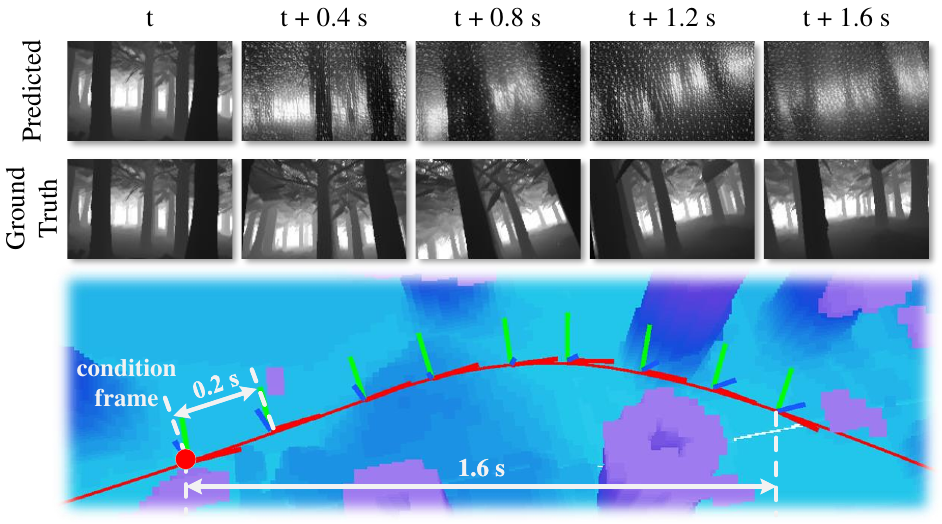}
    \caption{\textbf{Future-depth rollout predicted by FlowPilot.}
    The rollout preserves obstacle layout and ego-motion trend over the action horizon.}
    \label{fig:future_depth_prediction}
\end{figure}

Fig.~\ref{fig:future_depth_prediction} provides qualitative support: the rollout gives the action stream a temporally aligned estimate of upcoming free space. Freezing these tokens makes the policy more reactive and reduces both safety and speed, indicating that synchronous depth denoising acts as action guidance rather than only an auxiliary prediction task.

To isolate trajectory reconstruction, the waypoint baseline uses the same $15$-dimensional output, training procedure, controller, and test scenes as Bernstein. Its five position waypoints are converted into a $C^2$-continuous reference using fixed finite-difference derivatives and quintic Hermite interpolation, without additional learnable parameters. Table~\ref{tab:tracking_ablation} shows that Bernstein achieves lower position tracking loss, position RMSE, and velocity RMSE. 

We additionally evaluate the sensitivity to Euler denoising steps and prediction horizon using offline trajectory-prediction errors on an independent test set. Table~\ref{tab:offline_mae} reports the position, velocity, acceleration, and jerk MAE under these settings. The two-step configuration gives the lowest MAE across the evaluated denoising-step settings, supporting its use as the default inference configuration.

\begin{table}[!t]
    \centering
    \caption{Bernstein versus Waypoint tracking errors.}
    \label{tab:tracking_ablation}
    \footnotesize
    
    \begin{tabular}{lcc}
    \toprule
    Metric & Bernstein & Waypoint \\
    \midrule
    Position tracking loss (m$^2$)
        & $\boldsymbol{0.020}$ & $0.037$ \\
    Position RMSE (m)
        & $\boldsymbol{0.140}$ & $0.192$ \\
    Velocity RMSE (m/s)
        & $\boldsymbol{0.098}$ & $0.213$ \\
    \bottomrule
    \end{tabular}
\end{table}

\begin{table}[t]
    \centering
    \caption{MAE ablation of denoising steps and prediction horizons.}
    \label{tab:offline_mae}
    \footnotesize
    
    \resizebox{\columnwidth}{!}{
    \begin{tabular}{lcccccc}
    \toprule
    \multicolumn{1}{c}{Metric}
    & \multicolumn{1}{c}{\cellcolor{gray!15}Default}
    & \multicolumn{3}{c}{Euler denoising steps}
    & \multicolumn{2}{c}{Prediction horizon (s)} \\
    \cmidrule(lr){3-5}\cmidrule(lr){6-7}
    & & 3 & 4 & 5 & 0.8 & 2.4 \\
    \midrule
    Position MAE (m)
    & \cellcolor{gray!15}$\boldsymbol{0.132}$
    & $0.137$ & $0.139$ & $0.141$ & $\boldsymbol{0.021}$ & $0.277$ \\
    Velocity MAE (m/s)
    & \cellcolor{gray!15}$\boldsymbol{0.294}$
    & $0.303$ & $0.307$ & $0.311$ & $\boldsymbol{0.105}$ & $0.388$ \\
    Acceleration MAE (m/s$^2$)
    & \cellcolor{gray!15}$\boldsymbol{0.668}$
    & $0.683$ & $0.694$ & $0.709$ & $\boldsymbol{0.489}$ & $0.812$ \\
    Jerk MAE (m/s$^3$)
    & \cellcolor{gray!15}$\boldsymbol{2.644}$
    & $2.672$ & $2.712$ & $2.779$ & $3.254$ & $2.926$ \\
    \bottomrule
    \end{tabular}}
    
\end{table}

\subsection{Real-World Flight Experiments}
\label{subsec:realworld_experiment}

The real-world experiments test whether the simulated behavior remains feasible under onboard latency, sensor limits, and unstructured geometry. We first measure runtime on the Jetson Orin NX. The deployed stack runs FlowPilot through TensorRT and reconstructs Bernstein trajectories analytically on the host. As reported in Table~\ref{tab:timing}, FlowPilot's total inference latency averages $10.234$\,ms, dominated by the two-step flow denoising ($9.246$\,ms), with depth preprocessing measured separately and excluded from this total. Although YOPO attains a lower inference latency of $0.862$\,ms through one-shot prediction, FlowPilot jointly denoises future depth and trajectories, and this anticipatory guidance sustains agile flight at high speed and in dense clutter (Fig.~\ref{fig:sim_success_rate}), where purely reactive policies degrade. Meanwhile, the $10.234$\,ms latency permits replanning at roughly $100$\,Hz, satisfying the real-time requirement of onboard control. EGO-Planner, by contrast, spends $23.358$\,ms on mapping alone--already more than twice FlowPilot's total inference latency.

\begin{table}[!t]
    \centering
    \caption{Onboard latency breakdown.}
    \label{tab:timing}
    \scriptsize
    \begin{tabular}{lccc}
    \toprule
    Module (ms)& FlowPilot & YOPO & EGO-Planner \\
    \midrule
    Depth preprocessing & $0.817$ & $0.817$ & -\\
    Mapping & - & - & $23.358$\\
    Distilled depth encoder & $0.890$ & - & -\\
    Flow denoising, $2$ steps & $9.246$ & - & -\\
    Bernstein basis expansion & $0.087$ & - & -\\
    Total inference & $10.234$ & $0.862$ & -\\
    Planning & - & - & $0.771$\\
    \bottomrule
\end{tabular}
\end{table}

\begin{figure}[!t]
    \centering
    \includegraphics[width=0.95\columnwidth]{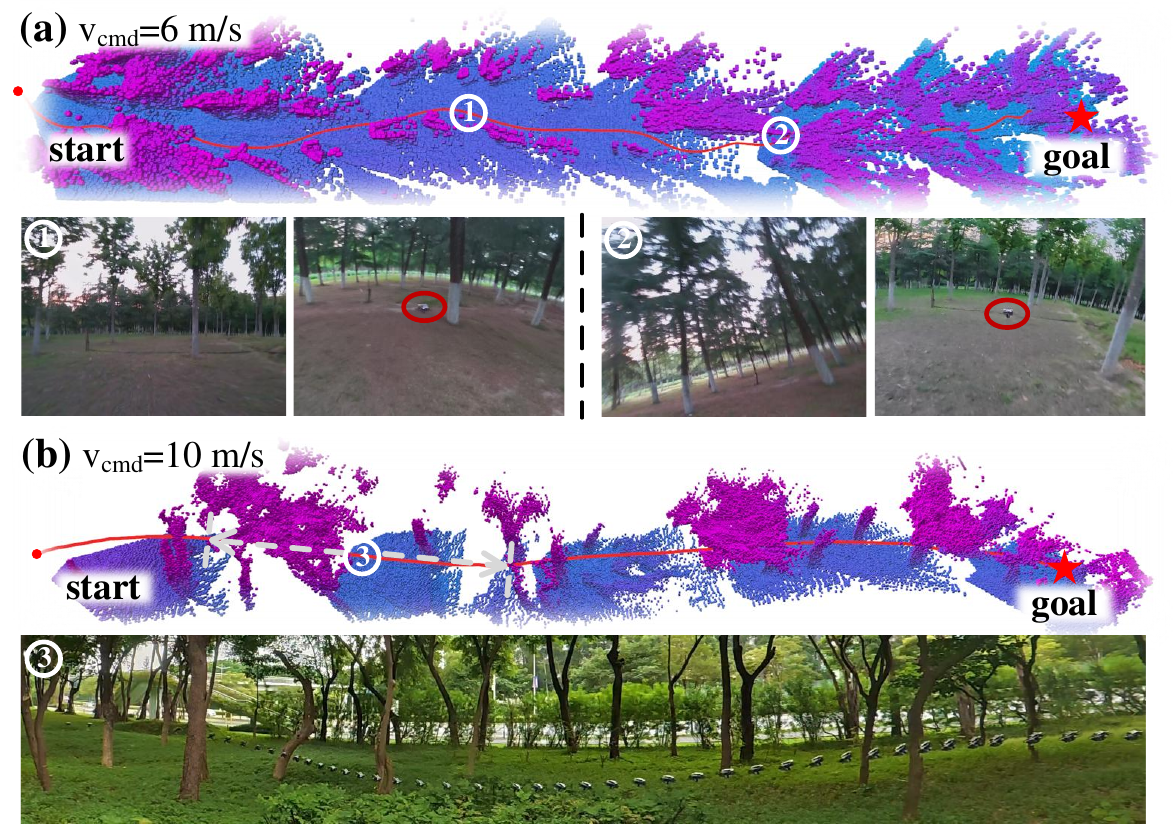}
    \caption{\textbf{Real-world outdoor flight experiments.} \textbf{(a)} A $100$\,m flight. \textbf{(b)} A $70$\,m flight. Each panel shows the flight path overlaid on a point-cloud map, with corresponding scene views at the numbered locations. The maps are used only for visualization.}
    \label{fig:forest_flights}
\end{figure}

Using this onboard stack, we evaluate FlowPilot in controlled indoor obstacle fields and natural outdoor forests. FlowPilot transfers zero-shot to the unseen layouts in Fig.~\ref{fig:platform_indoor}(b,c). In the shared layout (c), each tested speed uses $10$ trials (Fig.~\ref{fig:platform_indoor}(d)): FlowPilot maintains a $100\%$ success rate from $1$ to $6$\,m/s, whereas EGO-Planner drops to $0$ by $3$\,m/s.
We further evaluate FlowPilot in two outdoor forest environments, as shown in Fig.~\ref{fig:forest_flights}. In Fig.~\ref{fig:forest_flights}(a), FlowPilot completes a $100$\,m flight at $v_{\mathrm{cmd}}=6$\,m/s, reaching a peak speed of $5.5$\,m/s and a mean speed of $3.4$\,m/s. In Fig.~\ref{fig:forest_flights}(b), it completes a $70$\,m flight at $v_{\mathrm{cmd}}=10$\,m/s, reaching a peak speed of $10$\,m/s and a mean speed of $6$\,m/s. These outdoor flights demonstrate onboard perception and closed-loop control in unstructured terrain under the evaluated conditions.

\section{Conclusions}

We presented FlowPilot, a WAM that brings the joint video-action paradigm to onboard UAV navigation. A single MoT couples a depth-video expert and an action expert through shared attention, so that anticipating the future scene and generating the trajectory are one flow-matching computation rather than two separated stages. The key to making this practical on a drone is the action representation: by predicting the free control points of a state-constrained Bernstein polynomial, FlowPilot outputs trajectories that are $C^2$-continuous and dynamically feasible by construction, which a flight controller can track directly and which keep the denoised action latent small enough for real-time inference.

Trained on a three-level depth pyramid spanning large-scale IsaacLab simulation, photorealistic Flightmare rendering, and real onboard flight, FlowPilot transfers to physical quadrotors without ground-station computation or offline mapping. It achieves higher closed-loop success rates than baselines under increasing obstacle density and commanded speed, and achieves a mean inference latency of $10.2$\,ms on a Jetson Orin NX, reaching a peak speed of $10$\,m/s during outdoor flight. However, FlowPilot remains limited to short-horizon, goal-directed navigation from a single depth camera, making it vulnerable to thin obstacles, transparent surfaces, and regions outside the depth camera’s reliable sensing range. Future work should extend the model with longer-horizon or memory-based prediction and complementary sensing modalities, moving toward more general onboard autonomy.

\bibliographystyle{IEEEtranS}
\bibliography{bib/references}
\end{document}